\documentclass{article}
\usepackage{arxiv}

\usepackage[utf8]{inputenc} 
\usepackage[T1]{fontenc}    
\usepackage{hyperref}       
\usepackage{url}            
\usepackage{booktabs}       
\usepackage{amsfonts}       
\usepackage{nicefrac}       
\usepackage{microtype}      
\usepackage{lipsum}		
\usepackage{graphicx}
\usepackage{natbib}
\usepackage{doi}
\usepackage{balance}
\usepackage{graphicx}
\usepackage{algorithm}
\usepackage{color}
\usepackage{siunitx}
\usepackage{booktabs}
\usepackage{etoolbox}
\usepackage{tabularx}
\usepackage{adjustbox}
\usepackage{array}
\usepackage{xparse}
\usepackage{algpseudocode}
\usepackage{caption}
\usepackage{subcaption}
\usepackage[acronym,smallcaps]{glossaries}

\usepackage{amsmath,amsthm,amsfonts}
\usepackage{bm}

\NewDocumentCommand{\rot}{O{30} O{1em} m}{\makebox[#2][l]{\rotatebox{#1}{#3}}}%

\newcommand{\batchSize}{b}
\newcommand{\nClasses}{C}
\newcommand{\uncFunc}{u}
\newcommand{\distLab}{dl}
\newcommand{\distUnlab}{du}
\newcommand{\learner}{\theta}

\newcommand{\nnSampleSize}{k}
\newcommand{\preSelectionIterations}{j}
\newcommand{\labeledSet}{\mathcal{L}}

\newcommand{\unlabeledSet}{\mathcal{U}}

\newcommand{\MDPPolicy}{\pi}
\newcommand{\MDPStateSet}{\mathcal{S}}
\newcommand{\MDPActionsSet}{\mathcal{A}}
\newcommand{\MDPRewardsSet}{\mathcal{R}}
\newcommand{\MDPSingleState}{s}
\newcommand{\MDPPreSelectedActions}{\bm{a}}
\newcommand{\MDPCurrentRewards}{\bm{r}}
\newcommand{\MDPSinglePreSelectedAction}{x}
\newcommand{\MDPSingleReward}{r}

\newcommand{\AmountOfSDs}{\alpha}
\newcommand{\AmountOfTrainingALCycles}{\tau}
\newcommand{\ALCycleTime}{t}

\newcommand{\unlabeledQuery}{Q}

\newcommand{\nnOutputVector}{\mathbb{O}}
\newcommand{\nnInputVector}{\mathbb{I}}
\newcommand{\nnInputEncodingFunction}{E}

\DeclareMathOperator*{\argmax}{argmax}

\newcommand{\ImitAL}{\textsc{ImitAL}}

\newacronym{ML}{ML}{Machine Learning}
\newacronym{IL}{IL}{Imitation Learning}
\newacronym{AL}{AL}{Active Learning}
\newacronym{RL}{RL}{Reinforcement Learning}
\newacronym{MDP}{MDP}{Markov Decision Problem}
\newacronym{NN}{NN}{Neural Network}
\newacronym{LC}{LC}{Uncertainty Least Confidence}
\newacronym{MM}{MM}{Uncertainty Max-Margin}
\newacronym{Ent}{Ent}{Uncertainty Entropy}
\newacronym{QBC}{QBC}{Query-by-committee}
\newacronym{EER}{EER}{Expected Error Reduction}
\newacronym{SPAL}{SPAL}{Self-Paced Active Learning}
\newacronym{GD}{GD}{Graph Density}
\newacronym{QUIRE}{QUIRE}{Querying Informative and Representative Examples}
\newacronym{BMDR}{BMDR}{Query Discriminative and Representative Samples for Batch Mode Active Learning}
\newacronym{BatchBALD}{BatchBALD}{Efficient and Diverse Batch Acquisition
for Deep Bayesian Active Learning}

\glsdisablehyper

\newcolumntype{C}[1]{>{\centering\arraybackslash}p{#1}}
\newcolumntype{R}[1]{>{\raggedleft\arraybackslash}p{#1}}
\newcolumntype{L}[1]{>{\raggedright\arraybackslash}p{#1}}

\MakeRobust{\Call}

\title{\ImitAL{}: Learned Active Learning Strategy on Synthetic Data}
\date{August 24, 2022}

\author{\href{https://orcid.org/1234-5678-9012}{\includegraphics[scale=0.06]{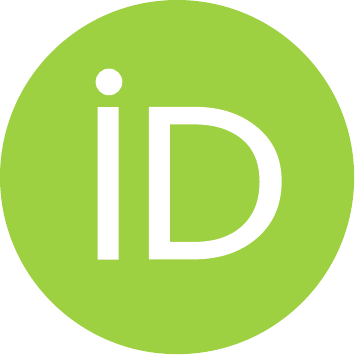}\hspace{1mm}Julius Gonsior\hspace{1mm}}\\
	Technische Universität Dresden\\
	Dresden, Germany\\
	\texttt{julius.gonsior@tu-dresden.de} \\
	\And
	\href{https://orcid.org/1234-5678-9012}{\includegraphics[scale=0.06]{orcid.pdf}\hspace{1mm}Maik Thiele\hspace{1mm}} \\
	Hochschule für Technik und Wirtschaft Dresden\\
	Dresden, Germany\\
	\texttt{maik.thiele@htw-dresden.de} \\
	\AND
	\href{https://orcid.org/0000-0001-8107-2775}{\includegraphics[scale=0.06]{orcid.pdf}\hspace{1mm}Wolfgang Lehner\hspace{1mm}}\\
	Technische Universität Dresden\\
	Dresden, Germany\\
	\texttt{wolfgang.lehner@tu-dresden.de} \\
}

\renewcommand{\shorttitle}{\ImitAL{}}

\hypersetup{
pdftitle={A template for the arxiv style},
pdfsubject={q-bio.NC, q-bio.QM},
pdfauthor={David S.~Hippocampus, Elias D.~Striatum},
pdfkeywords={First keyword, Second keyword, More},
}

\begin{document}
%

\maketitle              
\begin{abstract}
	\gls{AL} is a well-known standard method for efficiently obtaining annotated data by first labeling the samples that contain the most information based on a query strategy. In the past, a large variety of such query strategies has been proposed, with each generation of new strategies increasing the runtime and adding more complexity. However, to the best of our our knowledge, none of these strategies excels consistently over a large number of datasets from different application domains. Basically, most of the the existing \gls{AL} strategies are a combination of the two simple heuristics \textit{informativeness} and \textit{representativeness}, and the big differences lie in the combination of the often conflicting heuristics. Within this paper, we propose \ImitAL{}, a domain-independent novel query strategy, which encodes \gls{AL} as a learning-to-rank problem and learns an optimal combination between both heuristics. We train \ImitAL{} on large-scale simulated \gls{AL} runs on purely synthetic datasets.
	To show that \ImitAL{} was successfully trained, we perform an extensive evaluation comparing our strategy on 13 different datasets, from a wide range of domains, with 7 other query strategies.

\end{abstract}
\section{Introduction}
\gls{ML} has found applications across a wide range of domains and impacts (implicitly) nearly every aspect of nowaday's life.
Still, one of the most limiting factors of successful application of \gls{ML} is the absence of labels for a training set.
Usually, domain experts that are rare and costly are required to obtain a labeled dataset. Thus, to improve the manual label task is a prime object to improve.
For example, the average cost for the common label task of segmenting a single image reliably is 6,40 USD\footnote{According to scale.ai as of December 2021}.

Reducing the amount of necessary human input into the process of generating labeled training sets is of utmost importance to make \gls{ML} projects possible and scalable.
A standard approach to reduce the number of required labels without compromising the quality of the trained \gls{ML} model, is to exploit \acrfull{AL}. The approach consists of an iterative process of selecting exactly those unlabeled samples for labeling by the domain experts that benefit the to-be trained model the most. 
Given a small initial labeled dataset $\labeledSet = \{(x_i,y_i)\}_i^n$ of $n$ samples $x_i$ with the respective labels $y_i$ and a large unlabeled pool $\unlabeledSet = \{x_i\}, x_i \not\in \labeledSet$, an \gls{ML} model called \textit{learner} $\learner$ is trained on the labeled set.
A \emph{query strategy} then subsequently chooses a batch of $\batchSize$ unlabeled samples $\unlabeledQuery$, which will be labeled by the human experts and added to the set of labeled data $\labeledSet$. This \gls{AL} cycle repeats $\tau$ times until a stopping criterion is met.

By continuously retraining the model on the growing labeled set, one can avoid labeling samples with redundant information, which the model has already learned.
The challenge of applying \gls{AL} in this setting is the almost paradoxical problem to be solved: how to decide, which samples are most beneficial to the \gls{ML} model, without knowing the label of the samples, since this is exactly the task to be learned by the to-be-trained \gls{ML} model.

During the past years, many different \gls{AL} query strategies have been proposed, but to our knowledge, none excels consistently over a large number of datasets and from different application domains. 
For the popular application domains, e. g. natural language processing~\cite{schroder2020survey} and computer vision~\cite{beck2021effective}, it is indeed actively being researched which strategies to use, but for niche-domains it is often unclear which \gls{AL} strategy to use.
By deliberately focusing on domain-independent \gls{AL} strategies we aim to shed some light onto this problem.
Even though various extensive general survey papers~\cite{settles_al_survey,deep_al_surley,IJCAI21_survey} exist, no clearly superior \gls{AL} strategy has been identified.
The results of the individual evaluations in papers with newly proposed \gls{AL} strategies suggest that current \gls{AL} strategies highly depend on the underlying dataset domain.
Even more interestingly, the na\"ive baseline of randomly selecting samples often achieves surprisingly competitive results~\cite{LAL-RL,Pang_single,ALIL,QUIRE,ALBL,BMDR}.

At its core, the vast majority of \gls{AL} strategies rely on the same set of two simple heuristics: \textit{informativeness} and \textit{representativeness}. The first favors samples that foremost improve the classification model, whereas the latter favors samples that represent the overall sample distribution in the feature vector space.
Most recent \gls{AL} strategies add more layers of complexity on top of the two heuristics in their purest form, often resulting in excessive runtimes. 
This renders many \gls{AL} strategies unusable in large-scale and interactively operating labeling projects, which are exactly those projects that would benefit the most from ``optimal'' learning strategies.
\begin{figure*}[t]
	\centering
	\includegraphics[width=\textwidth]{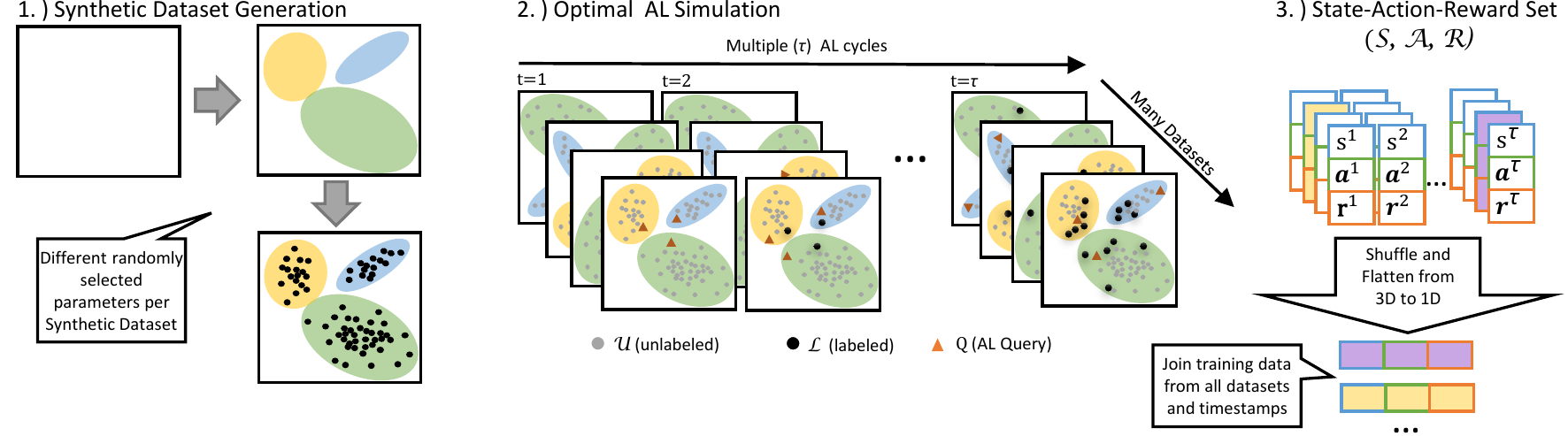}
    \caption{General overview on the training procedure of \ImitAL{}}
	\label{fig:SynD}
\end{figure*}

We are presenting \ImitAL{}, a  novel \gls{AL} strategy, which at its core is an \gls{NN} trained on very large simulated \gls{AL} episodes with the goal to optimally combine the basic \gls{AL} heuristics informativeness and representativeness. As it is not practically feasible to enumerate all possible real-world datasets as training data in the simulations, we are approximating them by using synthetic datasets instead. The benefit of synthetic datasets is that we can leverage the knowledge about all the labels to construct an optimal \gls{AL} strategy, which then serves as training basis for \ImitAL{}. The learning procedure of \ImitAL{} consists of \gls{IL}, where the training target is to imitate a demonstrated optimal strategy. 
Our work falls therefore under the category of ``learning \gls{AL} strategies''. According to our knowledge, our approach is, in contrast to similar works~\cite{LAL-RL,Pang_single,ALIL,ALBL}, the first one to solely utilize purely synthetical data to train the strategy. We can present a pre-trained, ready-to-apply \gls{AL} strategy which can be applied without any further necessary transfer-learning or fine-tuning in any domain.
The final version of \ImitAL{} is trained on 100,000 synthetic datasets with the respective true optimal \gls{AL} strategy simulations.

We start in Section~\ref{sec:Simulation} by presenting our synthetic datasets simulation process, followed by our \gls{IL} procedure in Section~\ref{sec:IL}. In Section~\ref{sec:evaluation}, we are comparing \ImitAL{} with 7 common \gls{AL} strategies on 13 real-world datasets. We are outperforming them with statistical significance on various evaluation metrics. We present the related work in  Section~\ref{sec:related_work} and conclude in Section~\ref{sec:conclusion}.
%
%
%
%
%
%
%
%
%
\section{Simulating \gls{AL} on Synthetic Training Data}
\label{sec:Simulation}
For the \gls{IL} training procedure of \ImitAL{} we need an \textit{expert} AL strategy, which the neural network behind \ImitAL{} can learn to imitate. In order to capture the characteristics of ``all'' possible datasets we pursue the idea by generating nearly ``infinite'' synthetic datasets and computing an optimal \gls{AL} strategy on them, leveraging the information about the known full labels for the synthetic datasets. 
%
\subsection{Synthetic Datasets}
\label{sec:SynD}
\begin{table}[t]
	\centering
	\begin{tabular}{lr}
		\toprule
		Parameter                 & Distribution and range \\
		\midrule
		\#samples                 & uniform(100, 5,000)    \\
		\#features                & uniform(2, 100)        \\
		\#classes                 & uniform(2, 10)         \\
		\#clusters per class      & uniform(1, 10)         \\
		\%class weights per class & dirichlet(0, 100)      \\
		\% of random label noise  & pareto(0, 100)         \\
		class separability        & uniform(0,10)          \\
		\bottomrule
	\end{tabular}
	\caption{Generation parameters for synthetic datasets}
	\label{tab:synthDatasets}
\end{table}
To be widely usable, \ImitAL{} should be applicable to every possible dataset and not be limited to a fixed domain or specific dataset characteristics.
Therefore, in an ideal setting, one would first enumerate all possible real-world datasets, run optimal \gls{AL} simulations on them, and train \ImitAL{} on the thereby observed state of the \gls{AL} problem.
Since it is practically and computationally not feasible to enumerate all real-world datasets, we make the assumption of being able to approximate them with a large enough number of random and diverse synthetic datasets and eventually achieve near-universal applicability of \ImitAL{}.
Obviously, the quality of \ImitAL{} inherently depends on the ``quality'' of the synthetic datasets; the more diverse and the more similar to real datasets the synthetic ones are, the more applicable \ImitAL{} will be in real-world usage scenarios.
For generating the synthetic datasets we use the implementation of the algorithm by~\cite{guyon2003design} in \emph{scikit-learn}~\cite{scikit-learn}, which is a runtime efficient method for creating a diverse range of synthetic datasets of varying shape and resulting classification hardness.
The general idea is to first generate clusters and then fill the clusters randomly with samples. The clusters are assigned for target classes to construct a supervised dataset. 
The clusters are placed either on the edges of a hypercube or a random polytope.
Table~\ref{tab:synthDatasets} lists the range of parameters and random distributions used while generating the synthetic datasets. 
The length of the hypercube sides is $2^{class\ separability}$, meaning larger values of $class\ separability$ result in more spread out samples and therefore harder classification tasks.
The class weight parameter controls the ratio between the number of samples between the classes.
We explicitly decided to not only focus on binary classification problems but to set the number of classes with a maximum of 10, as the more classes exist, the harder is the labeling process, and the more useful is \gls{AL} in general. Other parameter settings control the shape of the clusters, the distribution of the points in the clusters, the closeness of the cluster centers, random noise, the number of dimensions of the vector space, and many more characteristics.
We also did not decide to create very large datasets with many samples as this would firstly, drastically extend the training duration, and, secondly, \gls{AL} strategies should be able to easily scale up from small to large datasets.

Some rare parameter combinations result in very long runtimes. We included a timeout, after which we discarded the long-running parameter combination and tried out a different one.
This synthetic datasets generation approach has the benefit of being memory- and CPU efficient. A downside is that the implementation is limited to generate sample clusters using normal distribution. But as this is true for the vast majority of real-world datasets, we can accept this disadvantage.
%
\subsection{\acrlong{AL} Simulation and State-Action-Reward Set}
After a synthetic dataset has been generated, a simulation of the true optimal \gls{AL} strategy is performed on the dataset. 
As we know the true labels of all samples of the synthetic datasets in the \gls{AL} simulation, we can leverage this information to compute the true optimal \gls{AL} strategy.
We calculate the true optimal \gls{AL} decision on-the-fly during the simulation per \gls{AL} cycle by a \textit{future peek}: we compute for the current \gls{AL} cycle the resulting accuracy of the possible queries of unlabeled samples $\MDPSinglePreSelectedAction$, as if they each were added to the set of labeled samples $\labeledSet$. We will call this future accuracy in the following \textit{reward}.
We then choose the query $\unlabeledQuery$ with the highest resulting future accuracy and extend $\labeledSet$ by the newly labeled samples.
This cycle is performed $\AmountOfTrainingALCycles$-times, or until the set of unlabeled samples is empty.

As this process is computationally still quite heavy, we do not consider all possible batches, but perform a \textit{pre-selection} based on a heuristic, which selects a promising and diverse set of the top-$\nnSampleSize$ batches.
Details of the pre-selection are explained in Section~\ref{sec:IL}.

The results of the \gls{AL} simulation for each \gls{AL} cycle $\ALCycleTime$ for a specific synthetic dataset is a \textit{state-action-reward triple}. The state $\MDPSingleState$ is represented as a triple $\MDPSingleState =(\unlabeledSet^\ALCycleTime, \labeledSet^\ALCycleTime, \learner^\ALCycleTime)$, consisting of the set of unlabeled samples $\unlabeledSet^\ALCycleTime$, the set of labeled samples $\labeledSet^\ALCycleTime$, and the state of the learner model $\learner^\ALCycleTime$ trained on $\labeledSet^\ALCycleTime$. The corresponding actions $\MDPPreSelectedActions_{\MDPSingleState}$ is a set of the pre-selected queries $\MDPSinglePreSelectedAction$, whereas the respective rewards $\MDPCurrentRewards_{\MDPSingleState}$ for each of these actions is a set of rewards $\MDPSingleReward$. The optimal choice $\unlabeledQuery_{\MDPSingleState}^{\ALCycleTime} \in \MDPPreSelectedActions_{\MDPSingleState}$ for the \gls{AL} cycle $\ALCycleTime$ can be easily computed from the given accuracies -- the action with the highest future accuracy. Even though in theory the optimal choice would be sufficient as \gls{AL} strategy to learn from, the negative choices are also beneficial for the training of \ImitAL{} for practical reasons (Section~\ref{sec:IL}). We therefore preserve all respective rewards/accuracies in the state-action-reward triples.

This simulation is repeated $\AmountOfSDs$-times using different synthetic datasets. 
The accumulated state-action-reward pairs, denoted as the triple $(\MDPStateSet, \MDPActionsSet, \MDPRewardsSet)$, reflect then the input for \gls{IL} training procedure of the \gls{NN} of \ImitAL{}.
%
%
%
%
%
%
%
%
%
\section{Training a \glsentrylong{NN} by  \glsentrylong{IL}}
\label{sec:IL}
%
The final step of \ImitAL{} is to use the generated state-action-reward triples $(\MDPStateSet, \MDPActionsSet, \MDPRewardsSet)$ for training an \gls{NN} as \gls{AL} query strategy.
Therefore, we are deploying the \gls{ML} technique \gls{IL}~\cite{Michie94buildingsymbolic}, where demonstrated expert actions are being replicated by the \gls{ML} model. The training task for \ImitAL{} is to find patterns in the presented actions. As \glspl{NN} perform best when the task at hand can be solved by pattern recognition, we use an \gls{NN} as \gls{ML} model for \ImitAL{}. 

The resulting challenge is then to encode the state-action-reward triples into suitable input (state) and output (action) pairs for the \gls{NN}. 

Subsequently (Section~\ref{sec:MDP}) we will first redefine \gls{AL} as a \gls{MDP} followed by  the \gls{IL} learning process (Section~\ref{sec:il_algo}). Section~\ref{sec:state_space} and Section~\ref{sec:action_space} outline the policy network input and output encoding in detail.
%
%
%
\subsection{\acrlong{AL} as \acrlong{MDP}}
\label{sec:MDP}
In order to explain how to leverage \gls{IL} for training an \gls{AL} strategy, we are first redefining the \gls{AL} query selection as a \acrfull{MDP}. An \gls{MDP} is a set of states $\MDPStateSet$, a set of actions $\MDPPreSelectedActions$, and a real-valued reward function $\MDPCurrentRewards(\MDPSingleState, \MDPSinglePreSelectedAction)$ for taking action $\MDPSinglePreSelectedAction \in \MDPPreSelectedActions$ in state $\MDPSingleState$. Additionally, the \textit{markov property} must hold true, stating that the effects of an action depend only on the current state, not on the history of prior states.

We define \gls{AL} as an \gls{MDP} problem, where the state space consists of the labeled set $\labeledSet$, the unlabeled set $\unlabeledSet$, and the currently trained learner model $\learner$.
The action space consists of all possible unlabeled samples $\MDPSinglePreSelectedAction \in \unlabeledSet$.
The optimization goal of an \gls{MDP} is to find a \emph{policy} $\MDPPolicy: \MDPStateSet \mapsto \MDPActionsSet$ selecting the best action for a given state. For the case of \gls{AL}, that selection is done by the query strategy.
Instead of manually defining a policy we train an \gls{NN} to function as the policy for \ImitAL{}.
To construct a batch, one may select the top-$\batchSize$ actions $\MDPSinglePreSelectedAction$ to construct a query $\unlabeledQuery$. Basically, the \gls{NN} trained as \gls{AL} strategy / policy solves therefore the ranking problem of selecting the batch $\unlabeledQuery$ of the $\batchSize$-best unlabeled samples out of the set of unlabeled samples $\unlabeledSet$.
The input of the policy network will be the state and multiple actions, therefore, representing multiple samples to label, and the output denotes which of the possible actions to take, or which samples to label.
The \gls{AL} query strategy is implemented as an \gls{NN} solving a listwise learning-to-rank~\cite{Listwise} problem, ranking the set of unlabeled samples $\unlabeledSet$ to select the most informative batch $\unlabeledQuery$.
%
%
%
\subsection{Imitation Learning}
\label{sec:il_algo}
%
%
%
For training a policy in an \gls{MDP} setting two widely used concepts exist. The first one is \gls{RL}, where the policy network takes an action on its own, receives a reward, gets retrained and the whole cycle repeats itself. 
An alternative concept is \gls{IL}, where an expert demonstrates an optimal policy, which the corresponding policy network tries to replicate. For \gls{AL}, it is very easy to demonstrate optimal policy actions, the state-action-reward triples, representing the result of the simulated \gls{AL} episodes as outlined previously. The advantage of \gls{IL} over \gls{RL} is that the demonstrated policy can be gathered separately of the training step, which allows us to run the \gls{AL} simulations in a large-scale, highly parallelized cluster environment beforehand.

We use \emph{behavioral cloning}~\cite{Michie94buildingsymbolic} as a variant of \gls{IL} , which reduces \gls{IL} to a regular regression \gls{ML} problem. The desired outcome is a trained policy returning the optimal action for a given state. We use the state-action-reward set $(\MDPStateSet, \MDPActionsSet, \MDPRewardsSet)$ to extract an optimal policy $\hat{\MDPPolicy}$, which we are then demonstrating to the to-be-trained network: 
\begin{align}
    \hat{\MDPPolicy}(\MDPSingleState) = \argmax_{
    \MDPSinglePreSelectedAction \in \MDPPreSelectedActions_{\MDPSingleState} \allowbreak,
    \MDPSingleReward_\MDPSinglePreSelectedAction  \in \MDPCurrentRewards_{\MDPSingleState}
}{\left(\MDPSingleReward_\MDPSinglePreSelectedAction \right)}
\end{align}%
For a given state $\MDPSingleState$, the action set $\MDPActionsSet$ contains all pre-selected actions $\MDPPreSelectedActions_{\MDPSingleState}$ for this state; the reward set $\MDPRewardsSet$ contains the respective rewards $\MDPCurrentRewards_{\MDPSingleState}$.
As the optimal policy only contains the optimal actions, it can be used to construct the optimal batch by taking the $\batchSize$-highest actions.

Unfortunately, just demonstrating the optimal actions from $\unlabeledQuery$ based on the highest rewards requires a significant amount of training data until convergence of the training process can be reached. Therefore not only utilize the optimal actions, but also all the gathered rewards for all other actions as \textit{negative} examples. In other words: we train a policy network $\widehat{\MDPPolicy}$ predicting for a given state $\MDPSingleState$ and a possible action $\MDPSinglePreSelectedAction \in \MDPPreSelectedActions_{\MDPSingleState}$ -- which equals labeling the sample represented by this action -- the reward $\MDPSingleReward_\MDPSinglePreSelectedAction \in \MDPCurrentRewards_{\MDPSingleState}$. In our case, the expected future accuracy by demonstrating the true reward $\dot{\MDPSingleReward}$ function is:$
    \dot{\MDPSingleReward}(\MDPSingleState, \MDPSinglePreSelectedAction) = \MDPSingleReward_{\MDPSinglePreSelectedAction}
$.
The learned reward can then easily be used to construct the learned policy, which is what we need for \gls{IL}.
A common problem of the \gls{IL} variant behavioural cloning is the so-called \textit{distribution mismatch} that occurs when the demonstrated distribution of the state space is far too different from the distribution of the state space encountered in application, i.e. when the learned policy performs poorly in the beginning, and finds itself therefore in a previously not observed state -- because an optimal policy would never end up in that space -- and has therefore no training experience to rely on and has to guess the optimal action to take. But as we generated a very-large diverse state-action-reward set based on datasets of varying difficulty, it can be safely assumed that nearly all possible states are included in our training data.

%
%
%
\subsection{Policy \gls{NN} Input Encoding}
\label{sec:state_space}
%
%
%
%
Before using the state-action-reward set $(\MDPStateSet, \MDPActionsSet, \MDPRewardsSet)$ to train the policy network predicting the future accuracy, we first transform it into a fixed sized vector representation using feature encoding. 
\gls{NN} are limited by the number of the neurons to either a fixed size input, or when using \textit{recurrent \gls{NN}} to circumvent this limitation, they often suffer from the case of memory loss where the beginning of the input sequence is forgotten due to exploding or vanishing gradients~\cite{hochreiter2001gradient}. The last problem occurs more frequently the larger and more length-varying the input is, which is the case for \gls{AL}. 
The raw state $\MDPStateSet$ consists of triples $(\unlabeledSet, \labeledSet, \learner)$. Therefore, the length of the input layer does not only depend on the overall number of samples, but is also multiplied by the dimension of the feature space. We therefore transform the state-action-reward set before the training into a fixed size representation, additionally making \ImitAL{} dataset agnostic. The raw actions set $\MDPActionsSet$ may then contain -- depending on the number of unlabeled samples -- many possible samples, or just a few, varying again drastically. That underpins the already mentioned \textit{pre-selection} method, reducing the number of possible actions to a fixed size.

\begin{figure*}[t]
	\centering
		\includegraphics[width=\textwidth]{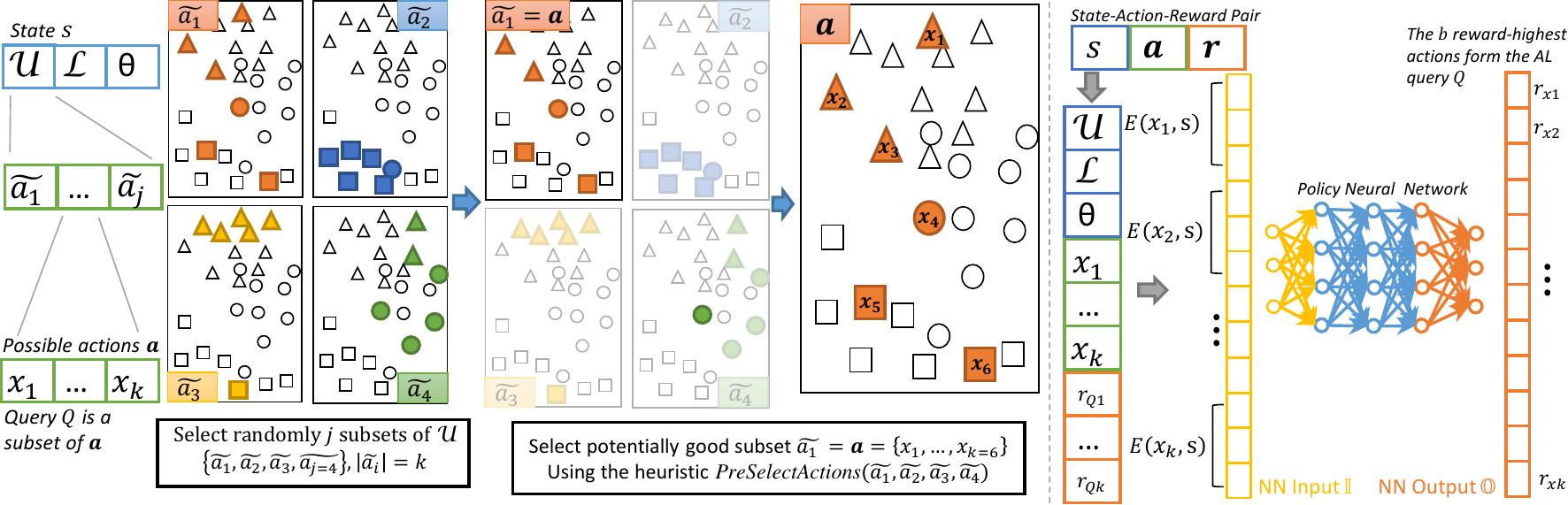}
		\caption{Pre-selection process and action meaning for \ImitAL{}, example for $\preSelectionIterations$= 4, $\nnSampleSize$=6, and  $\batchSize$=3, and encoding of a state-action-triple}
	\label{fig:single_state_encoding}
\end{figure*}
The transformation of the state-action-reward set into a suitable form for the policy network is called \textit{input} and \textit{output encoding}. Figure~\ref{fig:single_state_encoding} displays the general procedure of the encoding to the right. The input vector of the network is defined by $\nnInputVector$, the output vector by $\nnOutputVector$. The policy network should learn to calculate the expected reward $\widehat{\MDPSingleReward}_{\MDPSinglePreSelectedAction}$ for a given state $\MDPSingleState$ and a possible action $\MDPSinglePreSelectedAction$.
We chose a \textit{listwise} input encoding, where we enter $\nnSampleSize$ possible actions $\MDPSinglePreSelectedAction \in \MDPPreSelectedActions_{\MDPSingleState}, \lvert \MDPPreSelectedActions_{\MDPSingleState} \rvert = \nnSampleSize$ at once into the network, in contrast to a \textit{pointwise} encoding, where a single action is entered at a time. This has the benefit of enabling the network to compare each possible action relatively to the others, enabling \ImitAL{} therefore to take batch-aware \gls{AL} query decisions. The input of the policy network is defined by the vector $\nnInputVector_{\MDPSingleState}=\{\nnInputEncodingFunction(\MDPSinglePreSelectedAction, \MDPSingleState)|\MDPSinglePreSelectedAction \in \MDPPreSelectedActions_{\MDPSingleState}\}$, with $\nnInputEncodingFunction$ being the encoding of the action $\MDPSinglePreSelectedAction$. The output $\nnOutputVector = (\widehat{\MDPSingleReward}_1, \ldots, \widehat{\MDPSingleReward}_\nnSampleSize)$ of the network consists of exactly $\lvert \MDPCurrentRewards_{\MDPSingleState}  \rvert$ output neurons, one for each of the predicted accuracies $\widehat{\MDPSingleReward}_{\MDPSinglePreSelectedAction}$ for the respective possible actions $\MDPSinglePreSelectedAction$. The amount of output neurons equals therefore the amount of pre-selected actions: $\lvert \MDPCurrentRewards_{\MDPSingleState}  \rvert = \lvert \MDPPreSelectedActions_{\MDPSingleState} \rvert = \nnSampleSize$.

A single action represents an unlabeled sample $\MDPSinglePreSelectedAction \in \unlabeledSet$.
The input encoding function $\nnInputEncodingFunction(\MDPSinglePreSelectedAction, \MDPSingleState)$ defines on what basis the policy can make the \gls{AL} query strategy decision.
We use the state $\MDPSingleState=(\unlabeledSet, \labeledSet, \learner)$ to calculate the encoding, which is a 5-tuple consisting of multiple parts, the individual functions will be explained in the following paragraphs:
\begin{align}
\nnInputEncodingFunction(\MDPSinglePreSelectedAction, \MDPSingleState) =\left( \uncFunc_1(\MDPSinglePreSelectedAction, \learner),
\uncFunc_2(\MDPSinglePreSelectedAction, \learner), 
\uncFunc_3(\MDPSinglePreSelectedAction, \learner),
\distLab(\MDPSinglePreSelectedAction, \labeledSet), 
\distUnlab(\MDPSinglePreSelectedAction, \unlabeledSet)\right)
\end{align}%
The complete policy network input vector $\nnInputVector$ consists then of 5-times $\nnSampleSize$ values, an encoded 5-tuple for each unlabeled sample $\MDPSinglePreSelectedAction$ out of the set of possible actions $\MDPPreSelectedActions$: 
\begin{align}
   \nnInputVector_{\MDPSingleState} = \{\nnInputEncodingFunction(\MDPSinglePreSelectedAction_1, \MDPSingleState), \ldots, \nnInputEncodingFunction(\MDPSinglePreSelectedAction_\nnSampleSize, \MDPSingleState)\}, \MDPSinglePreSelectedAction \in \MDPPreSelectedActions_\MDPSingleState
\end{align}%
As mentioned in the beginning, a good \gls{AL} query strategy takes informativeness as well as representativeness into account.
Informativeness is derived by $\uncFunc_i(\MDPSinglePreSelectedAction, \learner)$, a function computing the uncertainty of the learner $\learner$ for the $i$-th most probable class for the sample $\MDPSinglePreSelectedAction\in\unlabeledSet$, given the probability of the learner $P_{\learner}(y|\MDPSinglePreSelectedAction)$ in classifying $\MDPSinglePreSelectedAction$ with the label $y$: 
\begin{align}
	\uncFunc_i(\MDPSinglePreSelectedAction, \learner) =
	\begin{cases}
		P_{\learner}\Bigl(\left(\argmax_{y,i} P_\learner(y|\MDPSinglePreSelectedAction)\right)\Bigm|\MDPSinglePreSelectedAction\Bigr), & \text{if } i \leq \nClasses \\
		0,                                                                         & \text{otherwise}
	\end{cases}
\end{align}%
$\argmax_{\_~,i}$ denotes the $i$-th maximum argument, and $\nClasses$ the number of classification classes.

For representativeness we compute $\distLab(\MDPSinglePreSelectedAction, \labeledSet)$ and $\distUnlab(\MDPSinglePreSelectedAction, \unlabeledSet)$, the first denoting the average distance to all labeled samples, the latter the average distance to all unlabeled samples:
\begin{align}
	\distLab(\MDPSinglePreSelectedAction, \labeledSet) &= \frac{1}{|\labeledSet|} \sum_{\MDPSinglePreSelectedAction_l \in \labeledSet} d(\MDPSinglePreSelectedAction,\MDPSinglePreSelectedAction_l), \quad
	\distUnlab(\MDPSinglePreSelectedAction, \unlabeledSet) = \frac{1}{|\unlabeledSet|} \sum_{\MDPSinglePreSelectedAction_u \in \MDPSinglePreSelectedAction} d(\MDPSinglePreSelectedAction,\MDPSinglePreSelectedAction_u),
\end{align}%
where $d(\MDPSinglePreSelectedAction_1,\MDPSinglePreSelectedAction_2)$ is an arbitrary distance metric between the points $\MDPSinglePreSelectedAction_1$ and $\MDPSinglePreSelectedAction_2$.
We use the Euclidean distance for small feature vector spaces, and recommend using the cosine distance for high-dimensional feature vector space.
%
%
%
\subsection{Policy \acrlong{NN} Output Encoding}
\label{sec:action_space}
%
%
%
The output encoding of the final layer of the policy network defines which action to take.
We use a final softmax layer of $\nnSampleSize$ output neurons, each per possible action. The $\batchSize$ highest output neurons indicate the samples for the unlabeled query $\unlabeledQuery$.
%
%
%
\subsection{Pre-Selection}
\label{sec:pre_selection}
%
%
%
Instead of considering all possible actions, we pre-select promising actions, whose individual samples have the largest diversity and whose individual samples are the furthest away from each other, similar to \cite{BatchBALD}. The pre-selection fulfills two objectives: first and foremost, we can present a fixed amount of actions to the policy network, and secondly it keeps the runtime of the simulations within a processable range. A positive side effect of the fixed-size input of the policy network is the low and static runtime of \ImitAL{}, which is almost independent of the size of the dataset. The effect is especially apparent with very large datasets.

We start the pre-selection by drawing randomly $\preSelectionIterations$ possible actions $\{\widetilde{a}_1, \dots, \widetilde{a}_\preSelectionIterations\}$, with each $\widetilde{a}$ being a subset of $\unlabeledSet$. After that we use a heuristic to select the top-$\nnSampleSize$ most promising actions $\MDPPreSelectedActions$ out of the random ones.
The pre-selection process is illustrated in Figure~\ref{fig:single_state_encoding} at the left side.

We are using a heuristic to filter out potentially uninteresting actions. By calculating the average distance to the already labeled samples of all the samples in each possible action set $\widetilde{a}$ and select the action set $\MDPPreSelectedActions$ having the highest average distance: $
	\MDPPreSelectedActions =\argmax_{\widetilde{a}} \sum_{\MDPSinglePreSelectedAction \in \widetilde{a}}\distLab(\MDPSinglePreSelectedAction, \labeledSet)$,
where $\widetilde{a}$ contains $\nnSampleSize$ unlabeled samples.
Thus, we are ensuring that we sample evenly distributed from each region in the sample vector space during the training process.
We compute the heuristic for $\preSelectionIterations$ random possible batches, instead of all possible subsets of $\unlabeledSet$.
%
%
%
%
%
%
%
\section{Evaluation}
\label{sec:evaluation}
%
%
%
The goal of \ImitAL{} is to learn a domain-independent \gls{AL} strategy from synthetic datasets, which combines the strength of both the basic informativeness and the representativeness heuristics. For evaluation, we are comparing \ImitAL{} therefore with 7 \gls{AL} strategies on 13 real-world datasets from varying domains.
%
\subsection{Experiment Details}
For evaluation we selected 13 datasets from the UCI \gls{ML} Repository~\cite{uci} with varying sample size, feature size, and application domain, similar to the evaluations of~\cite{Pang_single,LAL-RL,QUIRE,ALBL,BMDR}. As an additional larger dataset the table classification dataset DWTC~\cite{DWTC} was also included. For the experiments we started with a single random sample of each class, and ran the \gls{AL} loop with a batch size $\batchSize$ of 5 for 25 cycles, or until all data was labeled. We repeated this for 1,000 times with varying initial labeled samples to generate statistically stable results. As learner model $\learner$ a simple \gls{NN} with 2 hidden layers and 100 neurons each was used. The datasets were split randomly into a 50 \% train and 50 \% test evaluation set.

%
\begin{figure}[t]
	\centering
	\includegraphics[width=\linewidth]{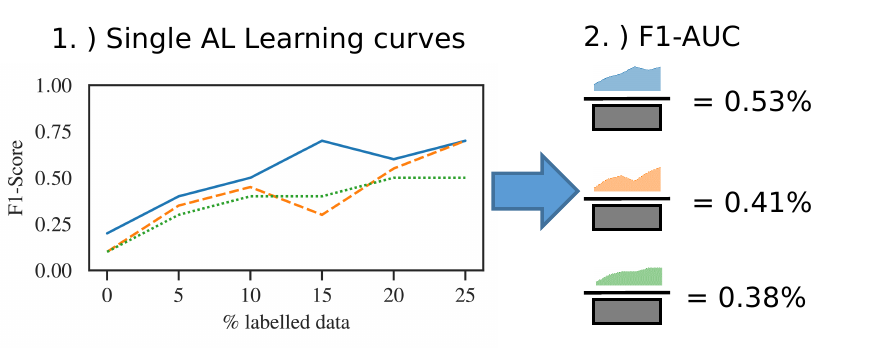}
    \caption{Converting learning curves into single numbers}
    \label{fig:f1auc}
\end{figure}
As evaluation metric we decided against \textit{learning curve plots} -- a standard \gls{ML} metric like accuracy or the F1-score shown for every \gls{AL} cycle -- and decided to use the \textit{area-under-the-curve} (AUC) of the learning curve instead, as has been also done recently in the \gls{AL} survey by Chan et. al.~\cite{IJCAI21_survey}. As we are repeating our experiments 1,000 times this decision makes it easier to compare many repeated experiments with each other. Similarly to~\cite{ALChallenge} we are further normalizing the AUC values by the maximum possible AUC value -- a rectangle of 100\% F1-Scores for each time step -- to additionally enable comparisons across datasets. As our evaluation datasets often have more than two classes, we decided to use the F1-Score as basic metric and denote our normalized AUC values as \textit{F1-AUC}. Figure~\ref{fig:f1auc} displays our method for computing the metric \textit{F1-AUC}. 

The training of \ImitAL{} is highly parallelizable, as the generation of the synthetic datasets and the respective \gls{AL} simulation may run completely in parallel. For a full training of \ImitAL{} with the best parameters we needed 100,000 computation jobs, resulting in a set of 1,000,000 state-action pairs as training data. In total, {\raise.17ex\hbox{$\scriptstyle\sim$}}1M CPU-hours were needed for all experiments conducted for this paper, including testing out different \gls{NN} and \gls{IL} configurations, and training the final version of \ImitAL{}. For the final version of \ImitAL{} we set the parameter of the simulated \gls{AL} cycle $\AmountOfTrainingALCycles$ to 10, the pre-sampling parameter $\nnSampleSize$ to 20 and $\preSelectionIterations$ to 10 during training, and 2 during application, as this suffices for a trained \ImitAL{}. The batch size was fixed to a standard value of 5 for the used UCI datasets.
\subsection{Comparison with other \acrlong{AL} strategies}
\label{sec:results}
\begin{table}[t]
    \addtolength{\tabcolsep}{2.7pt}
	\centering
	\begin{tabularx}{\linewidth}{lllllllll}
\toprule
            & \rot{ImitAL}   & \rot{LC}   & \rot{QBC}   & \rot{Ent}   & \rot{Rand}   & \rot{GD}   & \rot{BatchBALD}   & \rot{QUIRE}  \\
\midrule
 abalone    & 21.2 (2)        & 19.3 (5)    & 19.6 (4)     & 17.8 (6)     &\fontseries{b}\selectfont{21.3 (1)     }& 15.6 (7)    & 21.1 (3)           & 11.2 (8)      \\
 adult      &\fontseries{b}\selectfont{54.5 (1)       }& 53.5 (4)    & 54.1 (2)     & 53.5 (3)     & 51.8 (5)      & 47.9 (7)    & 51.3 (6)           &       \\
 australian &\fontseries{b}\selectfont{83.9 (1)       }& 83.8 (2)    & 83.8 (3)     & 83.8 (2)     & 83.0 (5)      & 83.6 (4)    & 79.8 (6)           & 71.5 (7)      \\
 BREAST     & 94.0 (3)        &\fontseries{b}\selectfont{94.4 (1)   }& 94.4 (2)     &\fontseries{b}\selectfont{94.4 (1)    }& 92.8 (4)      & 91.6 (5)    & 90.9 (6)           & 84.6 (7)      \\
 DWTC       &\fontseries{b}\selectfont{69.3 (1)       }& 65.3 (5)    & 65.9 (4)     & 63.4 (6)     & 67.8 (2)      & 52.8 (7)    & 66.1 (3)           & 50.1 (8)      \\
 fertility  & 88.2 (2)        & 87.8 (3)    & 87.7 (4)     & 87.8 (3)     & 87.0 (5)      &\fontseries{b}\selectfont{88.2 (1)   }& 86.8 (6)           & 86.8 (7)      \\
 flags      &\fontseries{b}\selectfont{57.5 (1)       }& 55.5 (6)    & 55.6 (5)     & 54.7 (7)     & 55.8 (4)      & 56.6 (2)    & 55.9 (3)           & 43.7 (8)      \\
 german     & 74.5 (2)        & 74.2 (4)    & 74.2 (5)     & 74.2 (4)     & 74.3 (3)      &\fontseries{b}\selectfont{75.5 (1)   }& 74.1 (6)           & 71.5 (7)      \\
 glass      &\fontseries{b}\selectfont{68.9 (1)       }& 67.6 (2)    & 67.4 (3)     & 66.6 (6)     & 67.3 (4)      & 66.3 (7)    & 67.1 (5)           & 40.6 (8)      \\
 heart      &\fontseries{b}\selectfont{79.0 (1)       }& 78.8 (3)    & 78.8 (5)     & 78.8 (3)     & 78.8 (4)      & 78.9 (2)    & 78.3 (6)           & 71.4 (7)      \\
 ionos &\fontseries{b}\selectfont{88.9 (1)       }& 88.6 (2)    & 88.5 (3)     & 88.6 (2)     & 88.0 (5)      & 88.2 (4)    & 82.9 (6)           & 53.5 (7)      \\
 wine       &\fontseries{b}\selectfont{95.2 (1)       }& 94.8 (2)    & 94.6 (4)     & 94.6 (3)     & 94.4 (5)      & 94.3 (6)    & 93.5 (7)           & 84.9 (8)      \\
 zoo        &\fontseries{b}\selectfont{93.7 (1)       }& 93.3 (2)    & 92.9 (6)     & 93.2 (3)     & 93.1 (4)      & 92.8 (7)    & 93.1 (5)           & 92.7 (8)      \\
 \midrule 
 mean \%    &\fontseries{b}\selectfont{74.5 (1)       }& 73.6 (3)    & 73.7 (2)     & 73.2 (5)     & 73.5 (4)      & 71.7 (7)    & 72.4 (6)           & 58.7 (8)      \\
 mean (r)   & 1.38             & 3.15         & 3.85          & 3.77          & 3.92           & 4.62         & 5.23                & 7.54           \\
\bottomrule
\end{tabularx}
    \caption{F1-AUC-scores (\%) for different \gls{AL} query strategies, mean for 1,000 repeated experiments each, including the ranks and the ranked mean. Empty cells indicate no calculable results within the maximum runtime window of seven days.}
	\label{tab:alipy_f1}
\end{table}
Our evaluation compares 7 \gls{AL} strategies against our \gls{AL} strategy, \ImitAL{}. The results are shown in Table~\ref{tab:alipy_f1}. Each displayed value is the mean of F1-AUC values for the 1,000 repeated runs. As the percentages are often quite similar, we additionally included the ranks. The displayed percentages are rounded, but the ranks are computed on the complete numbers, which can lead to different ranks for otherwise equally rounded displayed percentages.

We included \gls{LC} and \gls{Ent}~\cite{ent_sampling}, the two most common and basic variants of the informativeness heuristic, where greedily the most uncertain samples based on the classification probability of the learner model are selected for labeling. The \gls{GD} strategy~\cite{GD} was added as a pure representativeness heuristic-based strategy which solely focuses on sampling evenly from the vector space. BatchBALD~\cite{BatchBALD} is a popular \gls{AL} strategy which works well for computer vision deep neural networks. \gls{QUIRE}~\cite{QUIRE} is a computationally expensive combination of both heuristics, and \gls{QBC}~\cite{qbc_sampling} a combination of the uncertainty of multiple learner models~\cite{lc_sampling}\footnote{We used for all strategies the implementations from the open-source \gls{AL} framework ALiPy~\cite{ALiPy}}.

 \begin{figure}[t]
     \begin{center}
         \includegraphics{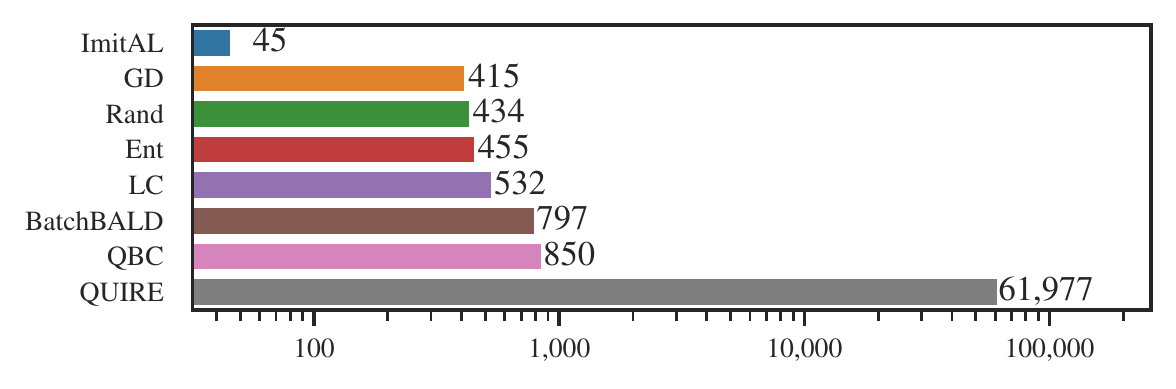}
         \caption{Average runtime duration in seconds per complete \gls{AL} experiment, with timeout duration for experiments lasting longer than seven days}
         \label{fig:performance}
     \end{center}
 \end{figure}
 
\ImitAL{} learns a combination of the two heuristics informativeness and representativeness. For the datasets \textsc{fertility}, \textsc{flag}, \textsc{german}, and \textsc{heart} \texttt{GD} is much better than \texttt{LC}. This is an indication that on these datasets a pure informativeness heuristic is challenged the most, whereas for the other strategies \texttt{LC} still seems to be the safest bet as a general-purpose \gls{AL} strategy. \ImitAL{} successfully learned to combine the best of both strategies, which can be especially seen by the superior performance on the datasets, where \texttt{GD} is better than \texttt{LC} and \ImitAL{} always managed to come close to the results of \texttt{GD}. \texttt{QBC} achieved quite competitive results, but at the cost of almost twice as high running cost than \ImitAL{} due to the expensive retraining of multiple learner models instead of a single one. The good results from the original \texttt{QUIRE} and \texttt{BatchBALD} paper could not be reproduced by us. Additionally, the runtime of \texttt{QUIRE} was so high that not even one \gls{AL} experiment finished within seven days. The pre-selection of \ImitAL{} with our used parameters means that \ImitAL{} always considers a fixed amount of 40 unlabeled samples during each \gls{AL} iteration, making it 10 times faster than even the second fastest \texttt{LC} strategy, which has to consider all unlabeled samples, as shown in Figure~\ref{fig:performance}. The benefit of \ImitAL{} of a constant runtime complexity becomes especially clear for the larger datasets. The displayed mean of the runtimes was calculated the same way as the F1-AUC-scores in Table~\ref{tab:alipy_f1}. Complex \gls{AL} strategies with high runtime are not practical in a real-world annotation scenario, as the annotation system can not afford to be unresponsive with background \gls{AL} calculations that take hours before the next sample can be labeled.

\begin{table}[t]
    \addtolength{\tabcolsep}{1.7pt}
	\centering
\begin{tabular}{lrrrrrr}
\toprule
 Dataset    & vs. QBC    & vs. LC     & vs. Ent    & vs. GD     & vs. BatchBALD   & vs. Rand   \\
\midrule
 BREAST     & 7  /43 /50 & 6  /46 /48 & 6  /46 /48 & 83 /16 /1  & 93 /6  /1       & 70 /25 /5  \\
 DWTC       & 72 /24 /4  & 76 /21 /4  & 86 /14 /1  & 100/0  /0  & 68 /26 /6       & 52 /36 /11 \\
 abalone    & 61 /22 /17 & 65 /13 /22 & 74 /22 /4  & 96 /4  /0  & 43 /26 /30      & 35 /30 /35 \\
 adult      & 39 /20 /41 & 41 /22 /38 & 42 /19 /39 & 70 /19 /11 & 59 /20 /21      & 53 /23 /24 \\
 australian & 18 /62 /20 & 20 /58 /21 & 20 /58 /21 & 34 /50 /16 & 82 /16 /2       & 46 /43 /11 \\
 fertility  & 8  /84 /7  & 7  /86 /7  & 7  /86 /7  & 11 /80 /9  & 19 /74 /6       & 16 /78 /6  \\
 flags      & 28 /65 /7  & 29 /64 /8  & 39 /55 /6  & 22 /63 /15 & 26 /63 /12      & 28 /61 /11 \\
 german     & 33 /38 /28 & 32 /40 /28 & 32 /40 /28 & 22 /38 /40 & 36 /33 /31      & 33 /36 /31 \\
 glass      & 28 /63 /8  & 25 /65 /11 & 39 /56 /5  & 49 /48 /4  & 40 /53 /7       & 36 /56 /8  \\
 heart      & 20 /67 /13 & 18 /69 /14 & 18 /69 /14 & 24 /55 /21 & 35 /49 /16      & 26 /54 /20 \\
 ionos      & 14 /72 /14 & 10 /72 /18 & 10 /72 /18 & 48 /46 /6  & 88 /12 /1       & 48 /47 /6  \\
 wine       & 9  /86 /5  & 5  /89 /6  & 10 /85 /5  & 30 /66 /4  & 47 /50 /3       & 30 /66 /4  \\
 zoo        & 5  /94 /1  & 3  /95 /2  & 3  /95 /2  & 17 /80 /3  & 14 /81 /5       & 12 /84 /4  \\
 Total      & 24 /59 /17 & 23 /60 /17 & 26 /58 /16 & 43 /46 /11 & 51 /40 /9       & 38 /50 /12 \\
\bottomrule
\end{tabular}
    \caption{Win-Tie-Losses in percentages across the repeated experiments based on Wilcoxon signed-rank test of \ImitAL{} compared to the six evaluation strategies}
    \label{tab:long_stats}
\end{table}
We also performed a significance test to prove that \ImitAL{} is not only by chance but indeed statistically sound better than the competitors. We used a Wilcoxon signed-rank test~\cite{Wilcoxon} with a confidence interval of 95\% to calculate the proportional win/tie/losses between \ImitAL{} and each competing strategy. For each of the 1,000 starting points, we took the F1-values of all the 25~\gls{AL} iterations~\footnote{As the exact p-values of the Wilcoxon signed-rank test are only computed for a sample size of up to 25, and for greater values an approximate -- in our case not existent -- normal distribution has to be assumed, we decided to stop our \gls{AL} experiments after 25 iterations.} of the two strategies to compare. Our null hypothesis is that the mean of both learning curves is identical. If the null hypothesis holds true we count this experiment repetition as a tie, and otherwise as a win or loss depending on which strategy performed according to the better mean. The results are displayed in Table~\ref{tab:long_stats}. Overall, \ImitAL{} won at least 35\% more often (in comparison to LC) compared to each strategy than lost against them. It also has to be noticed that the majority of the direct comparisons resulted in a tie. 
\section{Related Work}
%
%
%
\label{sec:related_work}
Besides traditional \gls{AL} strategies like~\cite{lc_sampling,mm_sampling,EER,GD,qbc_sampling,BMDR,SPAL} the field of learning \gls{AL} query strategies has emerged in the past few years.
Instead of relying on heuristics or solving optimization problems, they can learn new strategies based on training on already labeled datasets.

ALBL~\cite{ALBL} encodes in a \gls{QBC}-like approach the \gls{AL} query strategy as a multi-armed bandit problem, where they select between the four strategies \texttt{Rand}, \texttt{LC}, \texttt{QUIRE}, and a distance-based one.
To function properly, a reward for the given dataset is needed during the application of their approach as feedback.
In \cite{ALBL} a dedicated test set is used for feedback, which can in practice rarely be taken for granted.
Their tests on computing the reward on during \gls{AL} acquired labels showed that a training bias can occur resulting in poor performance.
 This only works well, when a separate test set is available, as the reward computed purely on the given labels during \gls{AL} can perform poorly because of training bias.
LAL~\cite{LAL} uses a random forest classifier as the learner, and statistics of the state of the inner decision trees from the learner for the unlabeled samples as input for their learned model.
Based on this they predict the expected error reduction.

Most of the other methods rely on \gls{RL} to train the \gls{AL} query strategy.
A general property that distinguishes the \gls{RL}  and imitation learning-based methods is the type of their learning-to-rank approach.
\cite{WoodwardandFinn,PAL,LAL-RL,ALIL} all use a pointwise approach, where their strategy gets as input one sample at a time. 
They all need to incorporate the current overall state of $\labeledSet$ and $\unlabeledSet$ into their input encodings to make the decision work on a per-sample basis.
\cite{LAL,Bachman,Pang_single}, as well as \ImitAL{}, use the listwise approach instead, where the strategy gets a list of unlabeled samples at once as input.
This also has the benefit of a batch-aware \gls{AL} setting.

The most distinctive characteristic of the \gls{RL}  and \gls{IL}-based approaches is the input encoding.
\cite{Bachman} uses the cosine similarities to $\labeledSet$ and $\unlabeledSet$ as well as the uncertainty of the learner.
\cite{ALIL} incorporates directly the feature vectors, the true labels for $\labeledSet$, and the predicted label of the current pointwise sample.
Both works also use imitation learning instead of \gls{RL}  with the same future roll-out of the \gls{AL} cycle as the expert as we propose for \ImitAL{}.
\cite{PAL} adds to their state additionally the uncertainty values for $\unlabeledSet$ and the feature vectors.
This has the limitation that the trained strategies only work on datasets having the same feature space as the datasets used in training.
\cite{Pang_single} bypasses this restriction by using an extra \gls{NN} which maps the feature space to a fixed size embedding.
Therefore they are, at the cost of complexity of an additional layer, independent of the feature space, but can still use the feature vectors in their vector space.
A big limitation is that their embedding currently only works for binary classification.
Additionally, they add distance and uncertainty information to their state.
\cite{LAL-RL} does not add the vector space into their input encoding.
To incorporate the current state for their pointwise approach, they use the learners' confidence on a fixed size set of unlabeled samples.
Further, they use the average distances to all samples from $\labeledSet$ and $\unlabeledSet$ each as well as the predicted class to encode the pointwise action.

Our input encoding uses uncertainty and distance.
Due to our listwise approach, we only need to add this information for the to-rank samples to our state, and not $\labeledSet$ or $\unlabeledSet$.
We explicitly do not add the feature vectors, as this limits the transferability of the trained strategy to new datasets, which contradicts our goal of a universal \gls{AL} strategy.

Most of the works rely on training on domain-related datasets before using their strategy.
This prevents an already trained \gls{AL} query strategy from being easily reusable on new domains.
\cite{LAL} bypasses this by training on simple synthetic datasets, but due to their simple nature, they still recommend training on real-world datasets too.
Our approach of using a large number of purely random and diverse synthetic datasets during training gives \ImitAL{} the benefit of not needing an explicit prior training phase on domain-related datasets.
The necessary pre-training of many related works as well as the often not publicly available code prevented them from being included in our evaluation.
%
%
%
%
%
%
%
%
%
\section{Conclusion}
%
%
%
\label{sec:conclusion}
We presented a novel approach of training a universally applicable \gls{AL} query strategy on purely synthetic datasets by encoding \gls{AL} as a listwise learning-to-rank problem.
For training, we chose \gls{IL}, as it is cheap to generate a huge amount of training data when relying on synthetic datasets.
Our evaluation showed that \ImitAL{} successfully learned to combine the two basic \gls{AL} heuristics informativeness and representativeness by outperforming both heuristics and other \gls{AL} strategies over multiple datasets of varying domains.
In the future, we want to include more requirements of large \gls{ML} projects into the state-encoding of \ImitAL{} to make it more applicable.
Large multi-label classification target hierarchies are often very hard to label but propose new challenges for \gls{AL}. 
Additionally, different label often have varying costs and should be treated accordingly by the \gls{AL} query strategy~\cite{Donmez}.
On a similar note, given many labels by multiple noisy sources like crowdsourcing relabeling using \gls{AL} becomes important~\cite{Zhao} which will be part of our future research.
%
%
%
%
%
%
%
%
%

%
%
%
\bibliographystyle{unsrtnat}
\bibliography{paper}
\end{document}